\documentclass[
]{ceurart}

\usepackage{listings}
\usepackage [english]{babel}
\usepackage [autostyle, english = american]{csquotes}
\MakeOuterQuote{"}
\begin{document}

\copyrightyear{2023}
\copyrightclause{Copyright for this paper by its authors.
  Use permitted under Creative Commons License Attribution 4.0
  International (CC BY-NC-SA 4.0).}

\conference{}

\title{Scaling Evidence-based Instructional Design Expertise through Large Language Models
}


\author[1]{Gautam Yadav}[%
orcid=0000-0002-9247-1920 ,
email=gyadav@andrew.cmu.edu,
]

\address[1]{Carnegie Mellon University,
  5000 Forbes Ave Pittsburgh PA 15213, United States}

\begin{abstract}
This paper presents a comprehensive exploration of leveraging Large Language Models (LLMs), specifically GPT-4, in the field of instructional design. With a focus on scaling evidence-based instructional design expertise, our research aims to bridge the gap between theoretical educational studies and practical implementation. We discuss the benefits and limitations of AI-driven content generation, emphasizing the necessity of human oversight in ensuring the quality of educational materials. This work is elucidated through two detailed case studies where we applied GPT-4 in creating complex higher-order assessments and active learning components for different courses. From our experiences, we provide best practices for effectively using LLMs in instructional design tasks, such as utilizing templates, fine-tuning, handling unexpected output, implementing LLM chains, citing references, evaluating output, creating rubrics, grading, and generating distractors. We also share our vision of a future recommendation system, where a customized GPT-4 extracts instructional design principles from educational studies and creates personalized, evidence-supported strategies for users' unique educational contexts. Our research contributes to understanding and optimally harnessing the potential of AI-driven language models in enhancing educational outcomes.
\end{abstract}


\maketitle

\section{Introduction}

The incorporation of large language models, such as GPT-4 \cite{openai2023gpt4}, in learning engineering offers a range of benefits, including the generation of personalized content, augmentation of existing learning materials, and support in evaluation processes. Despite its potential, GPT-4’s reliability can be inconsistent, particularly in complex subject areas, leading to potential inaccuracies and biases. To ensure high-quality learning experiences, a balanced approach combining AI-generated content with human oversight is essential. 

Our primary aim is to bridge the gap between empirical educational research and its practical implementation, focusing on utilizing Large Language Models (LLMs) to streamline evidence-based instructional design. This goal is underscored by two comprehensive case studies that illustrate the potential of our approach.

In addition to presenting these in-depth examinations, we also explore future trajectories and limitations inherent in this area of research. By drawing these outlines, we aspire to foster a deeper understanding of the judicious application of AI-driven language models such as GPT-4 in education. This understanding, in turn, can empower educators to optimize the use of these potent tools in their instructional endeavors.

\section{Prior Work}

The integration of AI-driven language models, like GPT-4, in education, presents numerous advantages, such as the capacity to produce tailored content, enhance existing learning materials, and offer support in evaluation processes \cite{learningright2023}. However, while GPT-4 can generate content that appears confident and precise, its reliability may be inconsistent, particularly in complicated subject areas. This can potentially result in incorrect or substandard content \cite{lehnert2023ai}. Additionally, biases in AI, originating from the training data and human decision-making, could influence the generated content, resulting in inaccuracies \cite{shen2023chatgpt}. Expert knowledge in specific fields is crucial in validating and maintaining the quality of AI-generated educational content. Thus, although GPT-4 provides several benefits, a balanced approach that combines AI-generated content with human supervision remains vital to ensure high-quality learning experiences.

Previous research involving large language models has explored their application in educational settings, such as the use of models like GPT-4 for generating questions or providing hints/explanations to students \cite{elkins2023useful, prihar2023comparing,pardos2023learning}. However, the current literature, to the best of our knowledge, does not extend beyond the creation of single-step open-ended or selected-response questions. Various research studies highlight the effectiveness of active learning strategies such as Predict-Explain-Observe-Explain (PEOE) \cite{chine2022scenario}, faded worked examples \cite{salden2010expertise}, and self-explanation \cite{nagashima2021using}, given certain boundary conditions. Despite their proven efficiency, these assessments are not universally employed due to the time and expertise required to construct them and the challenge of making evidence-based decisions on the optimal strategy to use.

In our work, we explored the application of GPT-4, an AI-powered language model, in the development and assessment of educational content. This exploration has revealed valuable insights into the potential benefits and challenges associated with using GPT-4 to automate the creation of higher-order assessments. By sharing our findings, we aim to provide a well-rounded perspective on our suggestions for optimally harnessing AI technology in educational settings.

\section{Case Studies}
This section details two case studies drawn from my experience as a Learning Engineer at Carnegie Mellon University, where I focused on enhancing courses by incorporating active learning components for students.

\subsection{Case Study 1: E-learning Design \& Principles}

The first case involved a course called E-learning Design \& Principles. The instructor's objective was to address the 30 instructional principles outlined in \cite{koedinger2013instructional}, basing their approach on the following learning objective:

Learning Objective: Deliver nuanced and evidence-based guidance regarding the effectiveness of selected instructional principles, considering boundary conditions in a given context.

The selected assessment strategy was a scenario-based predict-observe-explain (POE) method for each instructional principle. However, creating a single case study, specifically for the Worked Example principle (refer Appendix), demanded several days and multiple iterations. Once an example was finalized, we employed one-shot prompting to scale it for other instructional principles. Here is a sample prompt for the spatial contiguity principle:

\begin{enumerate}
      \item What are the boundary conditions of using spatial contiguity principle? cite references for these boundary conditions where authors have done a study to reach this conclusion based on data and evidence
      \item Create assessments in form of predict-explain-observe-explain scenarios for spatial contiguity principle out of [feed previous prompt output here as references]
      
I want to generate assessments in the form of predict-explain-observe-explain scenarios for explaining boundary conditions of when spatial contiguity principle is applicable based on EVIDENCE IN RESEARCH.

For every multiple-choice question and short answer, we want to generate feedback.

Can you start by giving a detailed study description followed by PEOE exercises for each of the references generated above and summary in the end of how these features interact with each other to make decision?

Let me write you an example of PEOE scenarios for boundary conditions using cited research studies in Worked Example principle then you try writing it for spatial contiguity principle: [feed Worked Example principle case study here as an example from as described in Appnedix]
    \end{enumerate}

After many iterations, we finalized these prompts, for example, one of the iteration involved emphasizing evidence from research to prevent the generation of hypothetical scenarios over constructing scenarios from studies in the cited papers. As shown in Figure 1, the outputs for each principle still required iterative cycles with the subject matter expert (instructor).
According to our estimation of what would have been required without the automation, the time for subsequent principles was reduced by more than half. This significant reduction is based on our usual turn-around cycle, where we manually find relevant references and create an initial draft.

\begin{figure}
  \centering
  \includegraphics[width=\linewidth]{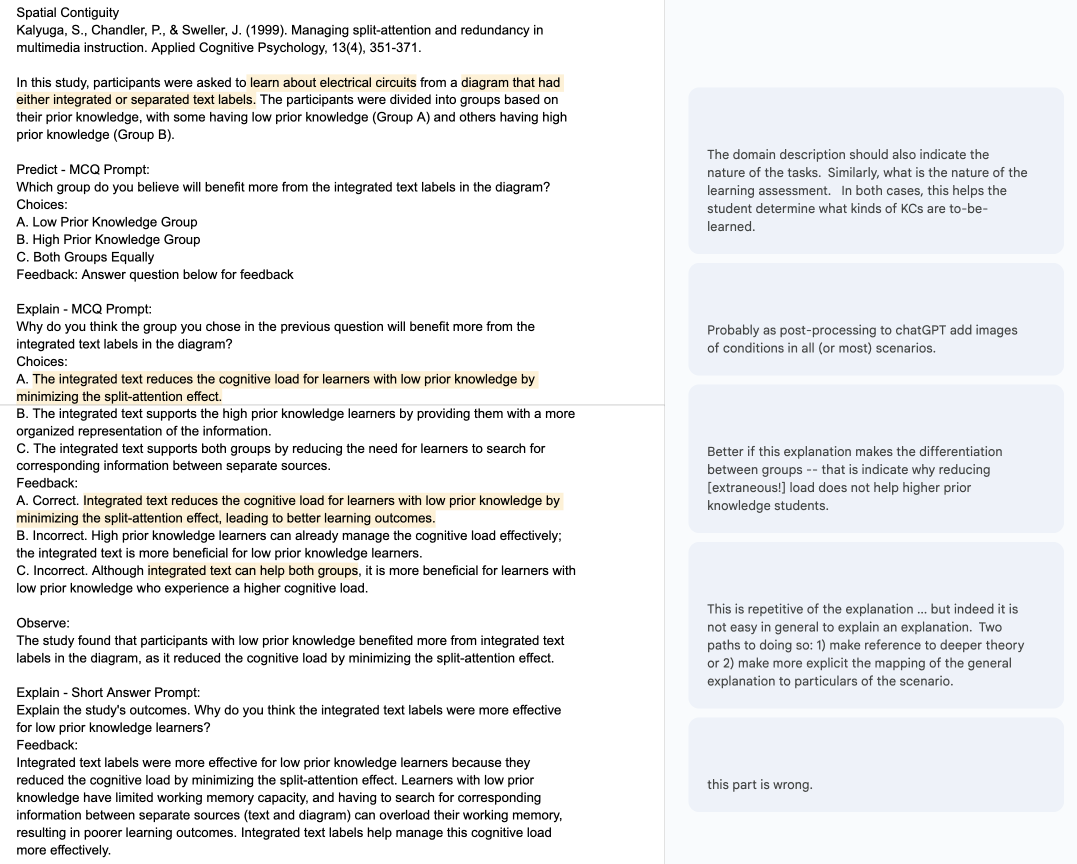}
  \caption{ Example of Predict-Explain-Observe-Explain activity created by GPT-4, focusing on the Spatial Contiguity Principle for one cited reference. A subject matter expert has added annotations to evaluate the quality of the content. The figure highlights both the strengths and limitations of using GPT-4 generated content in this context.
    }
    \label{fig:ui}
\end{figure}

\subsection{Case Study 2: Learning Analytics and Educational Data Science
}
The second case study pertains to a new course titled "Learning Analytics and Educational Data Science," slated for Fall 2023. There were no pre-existing online components, and the instructor wanted to develop programming 'learn-by-doing' assignments using Jupyter Notebook.

Learning Objective: Implement a predictive model using Python

Our chosen assessment strategy was the use of faded worked examples with feedback. We attempted to leverage data visualization problems developed using a combination of worked examples and problem-solving practice activities with feedback for another CMU course using GPT-4. However, these worked examples were inappropriate in this context, as students could simply copy-paste solutions so we only focused on crafting problem-solving activities.

We iteratively crafted a series of prompts, designed to yield the most effective output through trial and error:

\begin{enumerate}
      \item Can you give me 2 examples of hands-on exercises that cover the following learning objective “Implement a predictive model using Python” in a module called Classifiers.

      \item Can you provide a worked example in Python, including the corresponding code for the following hands-on exercise: [feed one example from the previous prompt output here]

      \item (3 - 5 Prompts):  [Debug any errors that appear when trying to run code provided in Prompt 2 output in Google Colab.; This average of 3 - 5 is based on the development of three exercises using different datasets for the learning objective above]

\item For each step in Jupyter Notebook,
    \begin{itemize}
        \item I want to create practice activities like these: 
        
[examples provided in the appendix]

Convert following code into above format where [code for this step] where students need to enter the given code with test cases to verify if students entered correctly.

\item (Only if a test case inadvertently revealed the answer in a previous step, we asked for more complex combinations to check for correct usage without giving away the answer) can you use more complex combinations to check for correct usage of above step without giving away the answer if students actually read the test cases

    \end{itemize}
    \end{enumerate}

We first attempted to use the same one-shot prompting strategy as in the previous case study but found that few-shot prompting yielded better results. Interestingly, GPT-4 did not generalize the Altair library in the output Python code as shown in Figure 2, even though all examples as shown in Appendix consisted of that. We only edited a few instructions where steps like reading the datasets or training classifiers were not suitable for this format and were covered in the student's prior knowledge.

\begin{figure}
  \centering
  \includegraphics[width=\linewidth]{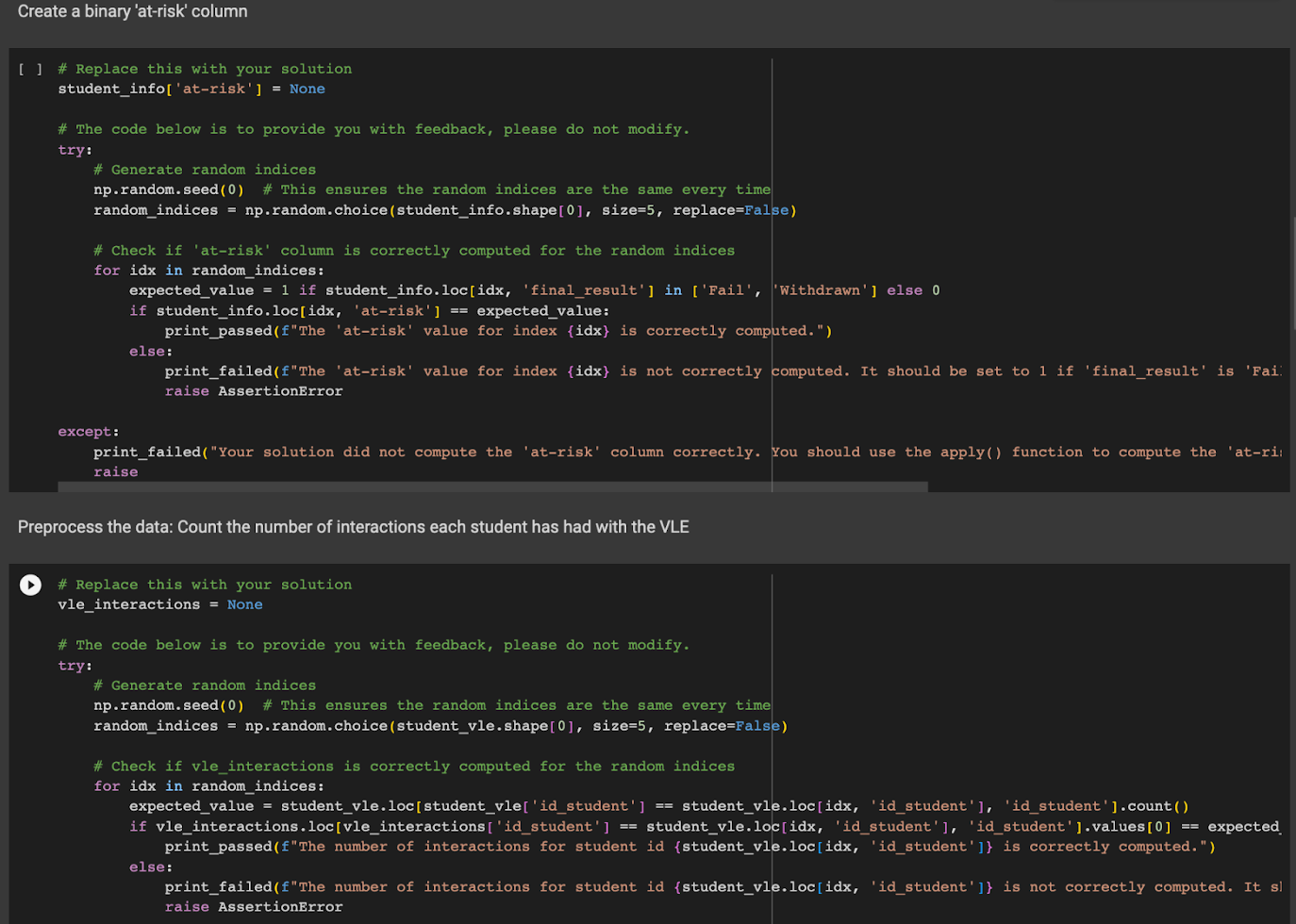}
  \caption{ A segment of a Jupyter Notebook showcasing a sequence of practice activities that were designed with the aid of GPT-4. The objective of these activities is to help students learn how to identify at-risk students using predictive models in Python.
    }
    \label{fig:ui}
\end{figure}

\section{Prompt Engineering - Best Practices for Instructional Design
}
Drawing from our experiences with GPT-4 in educational content creation, we have garnered invaluable insights into the potential advantages and obstacles of integrating AI in education. The lessons we've learned and their implications can guide educators and instructional designers to successfully implement AI-driven language models such as GPT-4, maximizing benefits while mitigating potential challenges. Here are our proposed best practices when utilizing large language models:

\subsection{Utilizing Templates for Instructional Design Tasks
}
Prior research \cite{reynolds2021prompt} indicates that as LLMs become more powerful, employing several examples (few-shot prompting) might not be as effective as zero examples (zero-shot prompting). Our case studies demonstrated the usefulness of examples in some prompts and zero-shot prompting in others. For complex tasks, like assessments involving specific instructional design principles, using well-defined examples in the form of templates can improve the quality of the generated output. Templates help enhance the structure and consistency of the materials generated by AI-driven language models, promoting a more streamlined content creation process.

\subsection{Fine-Tuning for Novel Instructional Problems
}
When creating new problems similar to the input's problem structure, a lower temperature value in the prompt can maintain a focus on the same knowledge components. Conversely, for diversity and problems in various contexts, such as a story or equation variable problems, a higher temperature can facilitate the broader transfer \cite{vertexai2023}.

\subsection{Handling Unexpected Output}
At times, LLMs may not behave as expected, even if all examples follow the same pattern. To counter this, consider a lower temperature for output similar to input or defining explicit rules in the prompt. Defining explicit rules needed involves iterating your prompt, identifying the patterns it struggles with, and explicitly stating these as rules in your prompt.

\subsection{Implementing LLM Chains for Multi-step Instructional Tasks
}
Although the surface quality of AI-generated educational content may be tempting, it's often best to break the task into the smallest subtasks initially. Once you have established the output quality, you can consider combining them, provided the interaction between them doesn't compromise the output. This approach, known as LLM chaining, helps improve output quality by avoiding the pitfalls of asking LLMs to handle multiple or nested tasks in one prompt \cite{wu2022ai}.

\subsection{Citing References in AI-Generated Instructional Materials}
Including credible sources or citations in AI-generated content is not only beneficial for enhancing its accuracy and credibility but also simplifies the process of verifying expert knowledge. During the first case study, the inclusion of verifiable references served as a tool for cross-checking the validity of both AI-generated content and Subject Matter Expert (SME) assertions. For instance, an expert initially identified an output from ChatGPT as incorrect, but upon cross-verification, the content was found to be accurate. This highlights that SMEs, while experienced, may not possess comprehensive knowledge in all aspects. Therefore, not only does referencing guard against the occasional fabrication of references by Language Models like GPT-4, it also provides a mechanism to identify and correct potential misconceptions held by SMEs. This approach ensures the creation of robust and reliable AI-generated instructional materials.

\subsection{Evaluating Output with AI Critique and SMEs}
AI critique can augment content quality as shown in Figure 3. However, content should be verified with Subject Matter Experts (SMEs) before incorporating it. Despite LLM's ability to critique and improve its own output, users should critically evaluate generated content.

\begin{figure}
  \centering
  \includegraphics[width=\linewidth]{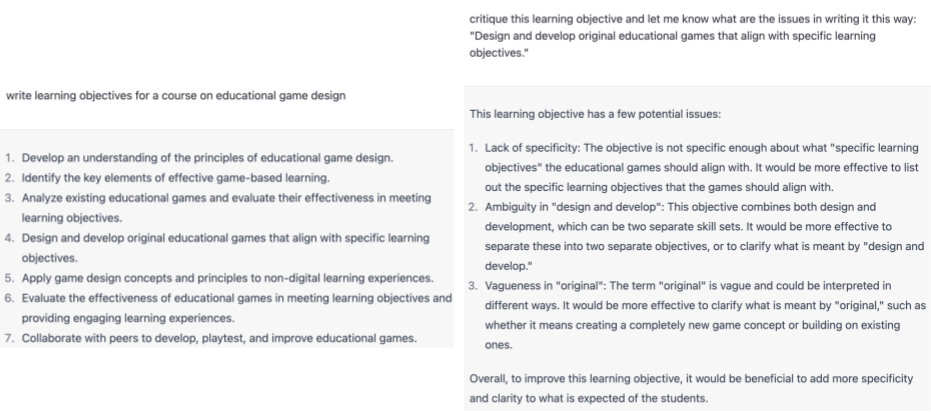}
  \caption{ChatGPT tasked to generate learning objectives for an educational game design, followed by critiquing one of them. The critique provided by ChatGPT gives meaningful recommendations for improvement, thereby demonstrating the model's self-awareness and its capacity to evaluate its own output.}
    \label{fig:ui}
\end{figure}

\subsection{Creating Rubrics and Grading}
AI can expedite rubric creation and grading, but educator involvement is crucial for reliability and fairness. Combined with a human educator, GPT-4 can enhance grading consistency and remove bias.

\section{Future Work}
In our future work, we aim to develop a sophisticated recommendation system, essentially "closing the loop" in our educational technology solution.

At the heart of our proposed system is a customized version of GPT-4, which we use to extract crucial information from empirical educational studies such as each paper's instructional design principles and identify the conditions under which they thrive. This extraction process enables us to encode and store data such as the educational domain, cognitive load, and learners' prior knowledge into a dedicated database, primed for future retrieval and application.

Harnessing the capabilities of GPT-4, we then create multiple assessment applications for each instructional design principle archived in our database. We adopt a few-shot prompting approach to devise these examples, aiming to guide users in effectively applying these principles across a broad range of educational contexts.

Our recommendation system is designed with the user's specific needs in mind. Users can input their unique instructional design requirements, including target learners, learning objectives, subject area, and other pertinent conditions. Our GPT-4 based system uses this information to generate evidence-supported instructional design strategies, tailored to the user's specific context. Each recommended strategy is paired with example applications and supported by original references from the studies they were drawn from, enabling users to further verify and delve into the source material.

The overarching goal of this endeavor is to democratize instructional design expertise, making it widely accessible to instructional designers and teachers alike. By doing so, we aim to streamline the design process, enhance educational outcomes, and ultimately drive forward the future of educational technology.

\begin{acknowledgments}
  I extend my sincere gratitude to Prof. Ken Koedinger and John Stamper, whose subject matter expert guidance was indispensable to the success of this research. Their profound wisdom and unwavering support enriched this work immeasurably. I also want to acknowledge students from Human Learning and How to Optimize It for their contribution to crafting the Model Human Learner document which inspired this idea.
\end{acknowledgments}

\bibliography{REFERENCES}

\appendix

\section{Worked Examples vs. Practice: A Content/Knowledge Demand Boundary Condition}

There were 4 research studies created for worked example principle to help students construct their own understanding of are boundary conditions of when to use worked examples instead of problem-solving along with the summary, here we present the first research paper only for reference on the format used for case study 1-

Kalyuga et al. (2001) Experiment 1 - Novices - Prediction:
Consider a study where participants were asked to write simple programmable logic controller (PLC) programs, and all participants had not been exposed to any training materials on relay circuits or PLC programming prior to the study. In one group (group A), participants alternated between 12 worked examples and 12 practice problems, whereas the other group (group B) received 24 practice problems.

Experiment 1 Description:
The procedure for the worked examples condition was identical to the problem-solving condition except that it included examples of relay circuits with corresponding steps in programming these circuits. Participants were requested to mentally follow all the steps according to a numbered sequence to ensure the programs were correct. The worked examples condition contained a series of 12 examples of relay circuits and programs for those circuits. Each example was followed by a problem-solving task. The 24 circuits used in this condition were identical to the circuits used in the problem-solving condition. The circuits with even numbers were problem-solving tasks identical for both treatments, while the circuits with uneven numbers were presented as worked examples in the worked example condition. Thus, in the worked example condition, participants studied 12 examples and attempted 12 problems, but in the problem-solving condition, participants attempted 24 problems. To avoid a split attention effect, separate program lines were embedded into the circuit with numbers of elements as closely as possible to the corresponding elements in the circuit. Participants could study each example as long as they wished.

\begin{table*}
  \label{tab:freq}
  \begin{tabular}{p{0.2\textwidth}p{0.7\textwidth}}
    \toprule
    Question Type & MCQ \\
    Prompt & Which group do you believe will exhibit better learning outcomes? \\
    Choices & \begin{enumerate}
        \item Worked Examples Group
\item Problem-Solving Group
\item No Difference
    \end{enumerate} \\
    Feedback & Answer question below for feedback \\
  \bottomrule
\end{tabular}
\end{table*}

\begin{table*}
  \label{tab:freq}
  \begin{tabular}{p{0.2\textwidth}p{0.7\textwidth}}
    \toprule
    Question Type & MCQ \\
    Prompt & Why do you think the group you chose in the previous question will exhibit better learning outcomes? \\
    Choices & \begin{enumerate}
        \item This group had less experienced learners who benefited from the reduced cognitive load of a worked example (Worked Example Effect).
\item This group had more experienced learners who benefited from the practice opportunity without the cognitive overhead of seeing a worked example (Redundancy Effect).
\item There should be no difference in the groups because each group experienced 24 learning opportunities.
    \end{enumerate} \\
    Feedback & \begin{enumerate}
        \item Correct. In their first experiment, the Worked Example Effect was found with inexperienced learners but disappeared after training (where more experienced students were exposed to training materials for a long enough time). See results for instructional efficiency below.
\item Incorrect. In their first experiment, the Worked Example Effect was found with inexperienced learners but disappeared after training (where more experienced students were exposed to training materials for a long enough time). See results for instructional efficiency below.
\item Incorrect. In their first experiment, the Worked Example Effect was found with inexperienced learners but disappeared after training (where more experienced students were exposed to training materials for a long enough time). See results for instructional efficiency below.
    \end{enumerate} \\
  \bottomrule
\end{tabular}
\end{table*}

Experiment 1 Outcomes: For these more novice students, worked examples produce greater instructional efficiency, that is, higher post-test question scores in less instruction time.

\begin{table*}
  \label{tab:freq}
  \begin{tabular}{p{0.2\textwidth}p{0.7\textwidth}}
    \toprule
    Question Type & Short Answer \\
    Prompt & Explain experiment 1 outcomes. Why do you think that the worked examples were more effective for this group? \\
    Feedback & According to Kalyuga et al. (2001): "For inexperienced learners, problem solving-based learning might overload the limited capacity of working memory, resulting in poor learning outcomes compared to a worked examples-based approach." \\
  \bottomrule
\end{tabular}
\end{table*}

\section{Data Visualization Problem-Solving Activities}
Figures 4 and 5 show examples of data visualization problems that were used for few-shot prompting in case study 2.
\begin{figure}
  \centering
  \includegraphics[width=\linewidth]{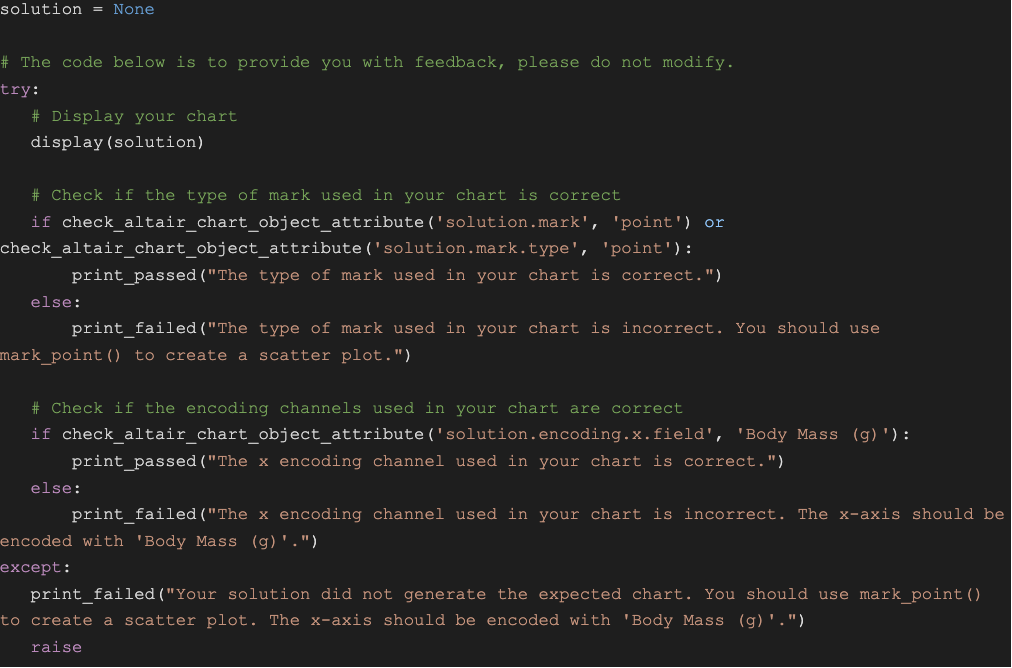}
  \caption{Example 1: Creating Scatter Plots in Altair}
    \label{fig:ui}
\end{figure}

\begin{figure}
  \centering
  \includegraphics[width=\linewidth]{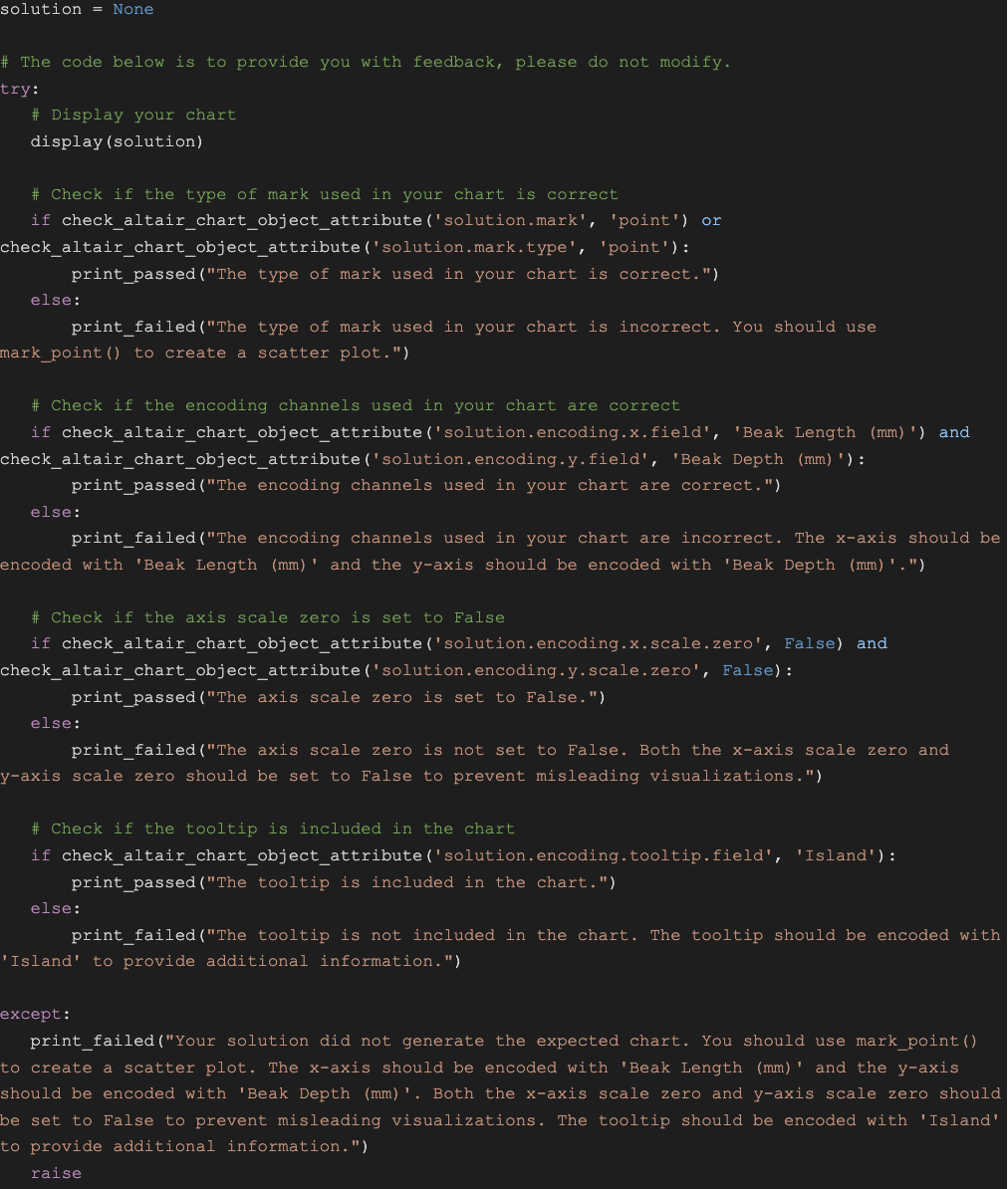}
  \caption{Example 2: Adding a Tooltip in Altair}
    \label{fig:ui}
\end{figure}

\end{document}